# Modeling and Simulation of a Point to Point Spherical Articulated Manipulator using Optimal Control

Prathamesh Saraf
Department of Electrical and Electronics Engineering
Birla Institute of Technology and Science – Pilani
Hyderabad, India
f20171348@hyderabad.bits-pilani.ac.in

Ponnalagu R. N.
Department of Electrical and Electronics Engineering
Birla Institute of Technology and Science – Pilani
Hyderabad, India
ponnalagu@hyderabad.bits-pilani.ac.in

*Abstract*— **This paper aims to design an optimal stability controller for a point to point trajectory tracking 3 degree of freedom (DoF) articulated manipulator. The Denavit Hartenberg (DH) convention is used to obtain the forward and inverse kinematics of the manipulator. The manipulator dynamics are formulated using the Lagrange Euler (LE) method to obtain a nonlinear system. The complicated nonlinear equations obtained are then linearized in order to implement the optimal linear quadratic regulator (LQR). The simulations are performed in MATLAB and Simulink and the optimal controller's performance is tested for various conditions and the results are presented. The results obtained prove the superiority of LQR over conventional PID control.**

## I. INTRODUCTION

Robotic manipulators are gaining tremendous popularity due to their high-quality performance and productivity in industrial applications [1]. The common applications of robotic manipulators in industries include object pick and place, painting, sketching, welding etc. All the said operations require basic position control strategies and can be adequately handled by the manipulators. Starting from simple single link manipulators, the flexibility and reachability of manipulators can go to much higher value based on the requirement of the application [2]. One of the common manipulator types is the 3 DoF spherical articulated manipulator and it is able to cover a complete 3-dimensional spherical volume as shown in Fig. 1.

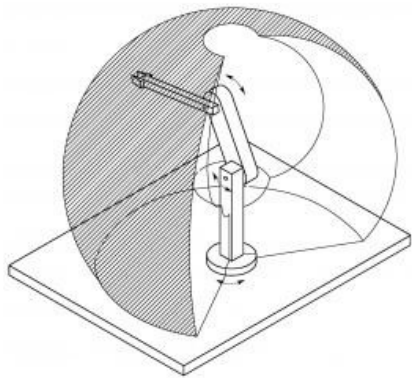

**Fig. 1:** Spherical workspace of the 3-DoF articulated manipulator

When the flexibility or DoF increase, the stability of the end effector will decrease which will reduce the performance and also make the system unstable. Vibrations make the control of such systems more challenging [3]. In order to improve the stability, controllers are used and the conventional one is the PID controller [4-7]. This classical control theory encompasses only the linear time-invariant single-input single-output systems, as the basis of control for these systems depends only on how their behaviour is modified using a feedback loop. On the other hand, LQR, a type of optimal control theory focuses on mathematical optimization of an objective cost function for a dynamic and multi-input multi-output systems, and is expected to be more robust for multi-link systems like the articulated manipulator. In such kind of systems as the number of joint parameters increase, the non-linearity and the system complexity increases and control becomes difficult. LQR focuses on non-linear models rather than the classical linear equation approach of PID. Also, in LQR control the system equations can be directly fed to the controller to acquire the desired response unlike the PID controllers' which requires linearization in every test and new condition.

Z. M. Doina [8] and Amit Kumar et al. [9] had explored LQR controllers' performance on a 2-link manipulator. Josias et al. [10] provided a comparative approach between PID and LQR controller for a cylindrical manipulator. Haleema et al. [11] solved the stability problem using pole placement technique for the 2 DoF manipulators. Mariam et al. [12] implemented the optimal controller for a single link manipulator and explained its performance. The literature shows that optimal controllers always provide better stability and accuracy for robotic manipulators. However, there has been major focus only on optimal control for 2 DoF manipulators. This motivated us to design and implement the full state feedback optimal controller for a 3 DoF spherical manipulator and test its performance for motion from one point to another. The manipulator specifications used for simulation are calculated based on torque requirement and are listed in Table I.

TABLE I.

| S. No. | Parameter | Value |
|---|---|---|
| 1. | Manipulator mass (*m*) | 2.5 Kg |
| 2. | Link 1 length (*a1*) | 25 cm |
| 3. | Link 2 length (*a2*) | 15 cm |
| 4. | Link 3 length (*a3*) | 15 cm |

Further sections of the paper are organized as follows. Section II details the kinematics of the spherical articulated manipulator. It also presents the dynamic equations governing its motion, linearization of the equation and development of state-space model. Section III provides a comprehensive study and formulation of the LQR controller. Section IV, the results and discussion section discuss the point to point motion and analysis using the LQR controllers and the performance is compared with the PID controller. Conclusions are presented in section V.

## II. MATHEMATICAL MODELING

### A. Denavit Hartenberg (DH) Convention

The dynamic formulation of any robotic system starts with plotting of DH parameters of the system. DH convention is used to determine the exact position of each joint parameter of the robotic system in space. It consists of projections concerning the previous joint coordinates. For a 3 DoF articulated manipulator, there will be four DH parameters (three joint parameters and one end effector) or three pairs. The articulated manipulator with the reference frame is shown in Fig. 2

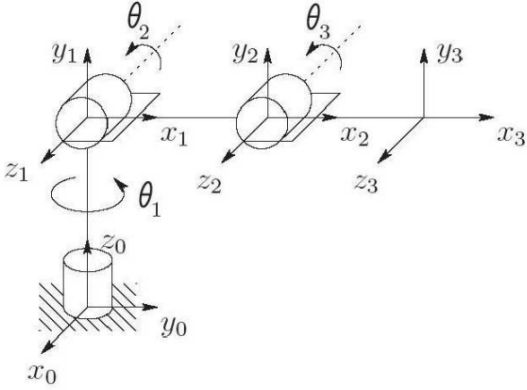

Fig. 2: The coordinate reference frame for each joint parameter of the manipulator

The DH parameters for the manipulator shown in Fig. 2 are listed in Table II:

TABLE II.

| $(i-1) - i$ | $a_i$ | $\alpha_i$ | $d_i$ | $\theta_i$ |
|---|---|---|---|---|
| 0 - 1 | 0 | 90° | $a_1$ | $\theta_1$ |
| 1 - 2 | $a_2$ | 0 | 0 | $\theta_2$ |
| 2 - 3 | $a_3$ | 0 | 0 | $\theta_3$ |

In Table II, $a$ is the length of the common normal, the distance between the $i^{th}$ and $i\text{-}1^{th}$ z-axis, $\alpha$ is the angle about the common normal between the $i\text{-}1^{th}$ z-axis and $i^{th}$ z-axis, $d$ is the distance between $i\text{-}1^{th}$ and $i^{th}$ x-axis, along $i\text{-}1^{th}$ z-axis and $\theta$ is the angle about the z-axis from the $i\text{-}1^{th}$ x-axis and the $i^{th}$ x-axis [14]. $\theta_1$, $\theta_2$, $\theta_3$ are the required joint parameters for the manipulator. $a_1$, $a_2$ and $a_3$ are the link lengths of the 3 links of the arm. The 4×4 transformation matrix is formed by the 3×3 rotation matrix and a 3×1 end point coordinate matrix of the current link is obtained using the DH parameter. The general form of transformation matrix is shown in Fig. 3. Using the general equation, the individual transformation matrix is obtained for each joint coordinate. The combined transformation matrix for the system is the sequential product of individual matrices and is presented in (1) and (2).

$$^{n-1}T_n = \begin{bmatrix} \cos\theta_n & -\sin\theta_n \cos\alpha_n & \sin\theta_n \sin\alpha_n & r_n \cos\theta_n \\ \sin\theta_n & \cos\theta_n \cos\alpha_n & -\cos\theta_n \sin\alpha_n & r_n \sin\theta_n \\ 0 & \sin\alpha_n & \cos\alpha_n & d_n \\ 0 & 0 & 0 & 1 \end{bmatrix} = \begin{bmatrix} R & T \\ 0\ 0\ 0 & 1 \end{bmatrix}$$

Fig. 3: General transformation matrix for DH convention

$$T_3^0 = T_1^0 T_2^1 T_3^2 \quad (1)$$

$$T_3^0 = \begin{bmatrix} C1C23 & -C1S23 & S1 & C1(a3C23 + a2C2) \\ S1C23 & -S1S23 & -C1 & S1(a3C23 + a2C2) \\ S23 & C23 & 0 & a3S23 + a2S2 + a1 \\ 0 & 0 & 0 & 1 \end{bmatrix} \quad (2)$$

where, $C_i = \cos\theta i$, $S_i = \sin\theta i$, $C_{ij} = \cos(\theta i + \theta j)$ and $S_{ij} = \sin(\theta i + \theta j)$

### B. Forward Kinematics

The forward kinematics is used to calculate the manipulator's position in space, provided the joint parameters are known for that position. In case of an articulated manipulator, the three joint angle values need to be specified in order to calculate the end effector coordinates in space. The forward kinematic equations for each link are presented in Table III:

TABLE III.

|  | i=0 | i=1 | i=2 | i=3 |
|---|---|---|---|---|
| $x_i$ | 0 | 0 | $a_2C_1C_2$ | $C_1*(a_3C_{12} + a_2C_2)$ |
| $y_i$ | 0 | 0 | $a_2S_1C_2$ | $S_1*(a_3C_{12} + a_2C_2)$ |
| $z_i$ | 0 | $a_1$ | $a_2S_2 + a_1$ | $a_3S_{23} + a_2S_2 + a_1$ |

### C. Inverse Kinematics

Inverse kinematics is the exact opposite of forward kinematics and uses the end effector position to calculate the joint parameters. The desired end effector position is specified and the equations are solved to calculate the joint angles in this case. This method is widely used in all robotic systems. Fig. 4 and Fig. 5 respectively show the side view and top view of the consideration for inverse kinematic equations and the equations (3) to (12) explain the inverse

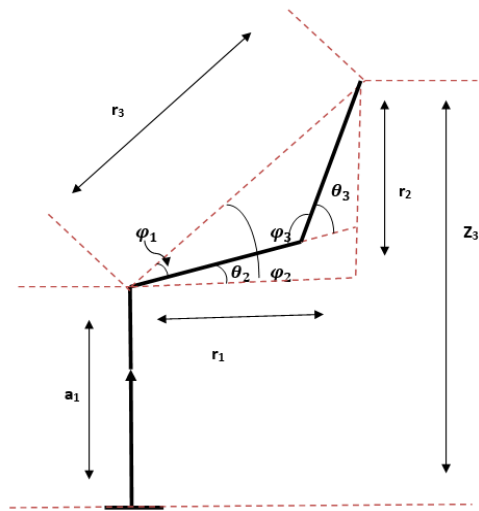

Fig. 4: Considerations for inverse kinematic calculations (side view)

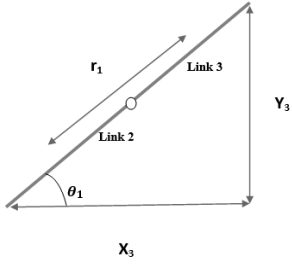

**Fig. 5:** Considerations for inverse kinematic calculations (top view)

kinematic calculations for the spherical articulated manipulator.

$$\theta_1 = tan^{-1}(\frac{Y_3}{X_3}) \quad (3)$$

$$r_1 = \sqrt{X_3^2 + Y_3^2} \quad (4)$$

$$r_2 = Z_3 - a_1 \quad (5)$$

$$\varphi_2 = tan^{-1}(\frac{r_2}{r_1}) \quad (6)$$

$$r_3 = \sqrt{r_1^2 + r_2^2} \quad (7)$$

$$\varphi_1 = cos^{-1}(a_3^2 - a_2^2 - r_3^2 / -2a_2 r_3) \quad (8)$$

$$\theta_2 = \varphi_2 - \varphi_1 \quad (9)$$

$$r_3^2 = a_2^2 + a_3^2 - 2a_2 a_3 \cos\varphi_2 \quad (10)$$

$$\varphi_3 = cos^{-1}(r_3^2 - a_2^2 - a_3^2 / -2a_2 a_3) \quad (11)$$

$$\theta_3 = \pi - \varphi_3 \quad (12)$$

where, $\varphi_1$, $\varphi_2$, $\varphi_3$ are the intermediate angles, $r_1$, $r_2$, $r_3$ are the intermediate side lengths, and $X_3$, $Y_3$, $Z_3$ are the end effector coordinates which will be user specified.

### D. Lagrangian Euler (LE) Formulation

LE formulation is the standard and fundamental method to obtain the second order dynamic equations of a robotic system. The Lagrangian equation (13) is defined as the sum of the difference between kinetic energy and potential energy of each joint parameter of the robot [14][15].

$$L = \sum_0^3 K_i - \sum_0^3 P_i \quad (13)$$

Once the Lagrangian equation is obtained, the equation of motion is determined as in (14):

$$\tau_i = \frac{d}{dt}\frac{\partial L}{\partial \dot{\theta}_i} - \frac{\partial L}{\partial \theta_i} \quad (14)$$

where, $i = 1, 2, 3$. $\tau$ is the torque required at each joint, $\theta$ and $\dot{\theta}$ are the angular positions and their respective angular velocities. The combined equation is obtained as in (15):

$$\tau = M(\theta)\ddot{\theta} + V(\theta,\dot{\theta}) + G(\theta) \quad (15)$$

In (15) $M(\theta)$ is the 3*3 inertia matrix, $V(\theta,\dot{\theta})$ is a 3*1 matrix consisting of the corriolis and centrifugal forces and $G(\theta)$ is the 3*1 gravity matrix. Equations (16) to (30) illustrates the three matrices for the articulate manipulator.

$$M(\theta) = \begin{bmatrix} M11 & M12 & M13 \\ M21 & M22 & M23 \\ M31 & M32 & M33 \end{bmatrix} \quad (16)$$

$$M11 = \frac{1}{2}m_1 a_1^2 + \frac{1}{2}m_1 a_2^2 + m_3(a_2^2\cos\theta_2^2 + \frac{1}{3}a_3^2\cos(\theta_2 + \theta_3)^2 + a_2 a_3\cos(\theta_2+\theta_3)\cos\theta_2) + \frac{1}{3}m_2 a_2^2 \cos\theta_2^2 \quad (17)$$

$$M12 = M21 = 0 \quad (18)$$

$$M13 = M31 = 0 \quad (19)$$

$$M22 = \frac{1}{3}a_2^2 m_2 + a_2^2 m_3 + \frac{1}{3}a_3^2 m_3 + a_2 a_3 m_3 \cos\theta_3 \quad (20)$$

$$M23 = M32 = \frac{1}{3}a_3^2 m_3 + a_2^2 m_3 + \frac{1}{3}a_2 a_3 m_3 \cos\theta_3 \quad (21)$$

$$M33 = \frac{1}{3}m_3 a_3^2 \quad (22)$$

$$V(\theta,\dot{\theta}) = \begin{bmatrix} V11 \\ V21 \\ V31 \end{bmatrix} \quad (23)$$

$$V11 = \left[-\frac{4}{3}m_2 a_2^2 \sin 2\theta_2 - \frac{1}{3}m_3 a_3^2 \sin 2(\theta_2 + \theta_3) - m_3 a_2 a_3 \sin(2\theta_2 + \theta_3)\right]\dot{\theta}_1\dot{\theta}_2 + \left[-\frac{1}{3}m_3 a_3^2 \sin 2(\theta_2 + \theta_3) - m_3 a_2 a_3 \cos\theta_2 \sin(\theta_2 + \theta_3)\right]\dot{\theta}_1\dot{\theta}_3 \quad (24)$$

$$V21 = [-m_3 a_2 a_3 \sin\theta_3]\dot{\theta}_2\dot{\theta}_3 + \left[-\frac{1}{2}m_3 a_2 a_3 \sin\theta_3\right]\dot{\theta}_3^2 + \left[\frac{1}{6}m_2 a_2^2 \sin 2\theta_2 + \frac{1}{6}m_3 a_3^2 \sin 2(\theta_2 + \theta_3) + \frac{1}{2}m_3 a_2^2 \sin 2\theta_2 + \frac{1}{2}m_3 a_2 a_3 \sin(2\theta_2 + \theta_3)\right]\dot{\theta}_1^2 \quad (25)$$

$$V31 = \frac{1}{2}m_3 a_2 a_3 \sin\theta_3 \dot{\theta}_2^2 + \left[\frac{1}{6}m_3 a_3^2 \sin 2(\theta_2 + \theta_3) + \frac{1}{2}m_3 a_2 a_3 \cos\theta_2 \sin(\theta_2 + \theta_3)\right]\dot{\theta}_1^2 \quad (26)$$

$$G(\theta) = \begin{bmatrix} G11 \\ G21 \\ G31 \end{bmatrix} \quad (27)$$

$$G11 = 0 \quad (28)$$

$$G21 = \frac{1}{2}m_3 g a_3 \cos(\theta_2 + \theta_3) + \frac{1}{2}m_2 g a_2 \cos\theta_2 + m_3 g a_2 \cos\theta_2 \quad (29)$$

$$G31 = \frac{1}{2}m_3 g a_3 \cos(\theta_2 + \theta_3) \quad (30)$$

### E. State Space Modeling

The dynamics formulated in the previous section are rearranged in a format which would consider the input signal and the initial conditions of the manipulator for computing, to perform the desired motion. The manipulator state-space system consists of six state variables, three input variables and three output variables. The state variables represent the absolute manipulator orientation in space consisting of three joint angles and their respective velocities. The input variables consist of the three torques of the three joints. The output consists of the required state variables for stability analysis of the spherical manipulator, which are the three joint angles. The state space equations for the manipulator are presented in (31) to (35):

$$\dot{X} = AX + BU \quad (31)$$

$$Y = CX + DU \quad (32)$$

$$X^T = [\theta_1 \quad \theta_2 \quad \theta_3 \quad \dot{\theta}_1 \quad \dot{\theta}_2 \quad \dot{\theta}_3] \quad (33)$$

$$U^T = [\tau_1 \quad \tau_2 \quad \tau_3] \quad (34)$$

$$Y^T = [\theta_1 \quad \theta_2 \quad \theta_3] \quad (35)$$

where A is a 6*6 state matrix, B is a 6*3 input matrix, C is a 3*6 output matrix, and D is a 3*3 null matrix as given in (36) and (37) [9-13][15]:

$$A = \begin{bmatrix} O_{3\times3} & I_{3\times3} \\ O_{3\times3} & -M(\theta)^{-1} \cdot [V(\theta,\dot{\theta}) + G(\theta)] \end{bmatrix}$$

$$B = \begin{bmatrix} O_{3\times3} \\ M(\theta)^{-1} \end{bmatrix} \quad (36)$$

$$C = \begin{bmatrix} 1 & 0 & 0 & 0 & 0 & 0 \\ 0 & 1 & 0 & 0 & 0 & 0 \\ 0 & 0 & 1 & 0 & 0 & 0 \end{bmatrix}; D = [O_{3\times3}] \quad (37)$$

Once the state-space equations are obtained, the optimal controller is designed for the articulated manipulator system. The following section describes the action of the optimal LQR controller on the manipulator system.

### III. LINEAR QUADRATIC REGULATOR

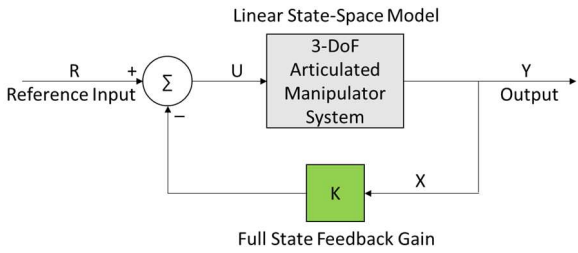

**Fig. 6:** The closed loop manipulator system with LQR full state feedback controller

#### A. Overview

The block diagram of the closed loop manipulator with LQR controller is shown in Fig. 6. The LQR controller involves a lot of mathematical computations to calculate the full state feedback matrix K. It uses an optimal control algorithm to minimize the cost function defined by the system equations. The cost function involves the state parameters and the input parameters of the system along with the Q and R matrices. The overall cost function needs to be minimum for optimal LQR solution. The Q and R matrices are the weights assigned to the state parameters and the input parameters. By varying the values of the two matrices, the total value of the cost function can be adjusted according to the desired output. The two main quantities that need to be optimized for the manipulator are the power consumption and response speed. For a faster response of the controller, the Q matrix values need to be high, whereas for minimizing the power consumption while achieving the desired setpoint without focusing on the time of response, the R matrix values need to be increased [16].

#### B. Equations

The LQR optimization problem requires a linearized state-space model of the system. The cost function which needs to be optimized is given in (38).

$$J = \int (X^T Q X + U^T R U) dt \quad (38)$$

After tuning the matrices according to the desired output matrix, the Q and R matrices are used to solve the Algebraic Riccati Equation (ARE) written in (39) to compute the full state feedback matrix, which is essentially the LQR controller.

$$A^T S + SA - SBR^{-1}B^T S + Q = 0 \quad (39)$$

The S matrix obtained from the ARE is used to calculate the full state feedback gain matrix K using the equation (40).

$$K = R^{-1} B^T S \quad (40)$$

The final control relation is then calculated as in (41).

$$U = -K * X \quad (41)$$

The tuned Q and R matrices which give the optimal response for the manipulator system parameters as defined before are given in (42) and (43).

$$Q = \begin{bmatrix} 10000 & 0 & 0 & 0 & 0 & 0 \\ 0 & 10000 & 0 & 0 & 0 & 0 \\ 0 & 0 & 10000 & 0 & 0 & 0 \\ 0 & 0 & 0 & 800 & 0 & 0 \\ 0 & 0 & 0 & 0 & 500 & 0 \\ 0 & 0 & 0 & 0 & 0 & 500 \end{bmatrix} \quad (42)$$

$$R = \begin{bmatrix} 1 & 0 & 0 \\ 0 & 1 & 0 \\ 0 & 0 & 1 \end{bmatrix} \quad (43)$$

The feedback gain matrix K computed is shown in (44).

$$K = \begin{bmatrix} 100 & 0 & 0 & 325.32 & 0 & 0 \\ 0 & 100 & 0 & 0 & 316.35 & 102.11 \\ 0 & 0 & 100 & 0 & 102.11 & 112.14 \end{bmatrix} \quad (44)$$

The next section presents the detailed comparative analysis of the LQR and PID controllers.

### IV. RESULTS AND DISCUSSION

The full state feedback LQR's performance is tested for various conditions on the articulated manipulator, and the results are presented in this section. The manipulator is fed in with the desired end-effector coordinates, and its motion is controlled from the start point until the desired endpoint. The system dynamics are programmed in MATLAB, and the controller is tested in Simulink. The manipulator's start position is fed in as the controller's initial conditions, and the goal position is the reference set point for the controller. The initial and final velocities of the three joint parameters are zero since the manipulator will start from rest and will be at rest once the set point is achieved. The initial and final points are user-based and arbitrarily chosen to test the complete performance. The test conditions are listed in Table IV.

TABLE IV.

| Cartesian co-ordinates | Value (cm) | Joint parameters | Value (rad) |
|---|---|---|---|
| $x_{initial}$ | 10 | $\theta_{1\,initial}$ | 0.7854 |
| $y_{initial}$ | 10 | $\theta_{2\,initial}$ | -1.6280 |
| $z_{initial}$ | 10 | $\theta_{3\,initial}$ | 1.6264 |
| $x_{final}$ | 15 | $\theta_{1\,final}$ | 1.0304 |
| $y_{final}$ | 25 | $\theta_{2\,final}$ | -0.3373 |
| $z_{final}$ | 20 | $\theta_{3\,final}$ | 0.3349 |

The controller is simulated in MATLAB for the initial and final conditions listed in Table IV and the response obtained for LQR controller and the PID controller are shown respectively in Fig. 7 and Fig. 8.

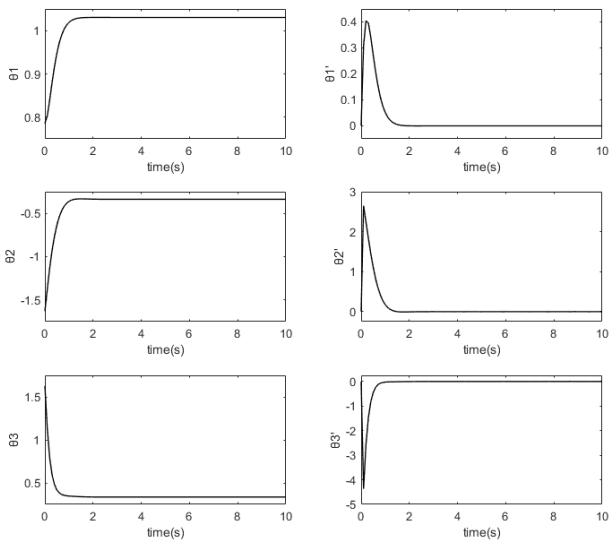

**Fig. 7:** Position and angular velocity response of the spherical manipulator's point to point motion for LQR controller

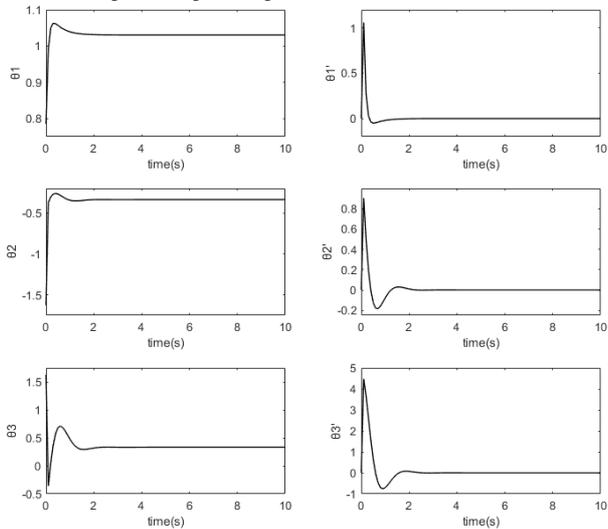

**Fig. 8:** Position and angular velocity response of the spherical manipulator's point to point motion for PID controller

It can be seen in Fig. 7, the curves start from the initial values and settle down at the final set point as mentioned in table 4. $\theta_1$ and $\theta_2$ takes 1.2 s to reach the setpoint while $\theta_3$ takes 1.5 s to settle. The velocity curves rise sharply from zero and settle down to zero once the angle setpoint is achieved. There is no undesired overshoot in any parameter response, which is the advantage of optiml controller. On the other hand in the response of PID controller [17] shown in Fig. 8 for the same conditions, the settling time for $\theta_1$ and $\theta_2$ is about 1.8 s while it is 2.1 s for $\theta_3$ which is more than the optimal controller response. The PID response shows overshoots before achieving the set point which is undesired and will create instability. Also, the velocity curves rise to higher values than those of LQR response which again is a sign of aggressive behaviour that is undesired.

## V. CONCLUSION

In this work the performance of a 3 DoF spherical articulated manipulator for a pick and place kind of operation is tested using the conventional PID controller and optimal LQR and the results are compared and presented. Although the PID controller is easy to implement, overshoots are present in the response curve which show an unstable behaviour. When the number of control variables increase, the number of PIDs need to be increased and tuning all of them may be a tedious task. Also, PID controller for a 3 DoF requires three feedback loops making the computation complex. As opposed to this, the LQR gives an optimal response and no overshoot is observed. Also, for an LQR control, only two matrices need to be tuned and only one control loop is required which reduces the complexity of the system. Hence an LQR controller is better suited for the spherical articulated manipulators' control mechanism than the classical PID controller in terms of output, complexity, and computation time.